\documentclass[letterpaper, 10 pt, conference]{ieeeconf}  % Comment this line out if you need a4paper

\IEEEoverridecommandlockouts                              % This command is only needed if 
\usepackage{graphicx}
\usepackage{xcolor}         % colors
\usepackage{amsmath}
\usepackage{caption}

\title{\LARGE \bf
Analysis of a Modular Autonomous Driving Architecture: \\ The Top Submission to CARLA Leaderboard 2.0 Challenge
}

\author{
Weize Zhang, Mohammed Elmahgiubi, Kasra Rezaee, Behzad Khamidehi, Hamidreza Mirkhani, \\ Fazel Arasteh, Chunlin Li, Muhammad Ahsan Kaleem, Eduardo R. Corral-Soto, Dhruv Sharma, \\
and Tongtong Cao
\vspace{2mm} \\
{\textit{Noah's Ark Lab, Huawei Technologies}} \\%
}

\begin{document}

\maketitle
\thispagestyle{empty}
\pagestyle{empty}

%%%%%%%%%%%%%%%%%%%%%%%%%%%%%%%%%%%%%%%%%%%%%%%%%%%%%%%%%%%%%%%%%%%%%%%%%%%%%%%%
\begin{abstract}

In this paper we present the architecture of the Kyber-E2E submission to the map track of CARLA Leaderboard 2.0 Autonomous Driving (AD) challenge 2023, which achieved first place. We employed a modular architecture for our solution consists of five main components: sensing, localization, perception, tracking/prediction, and planning/control. Our solution leverages state-of-the-art language-assisted perception models to help our planner perform more reliably in highly challenging traffic scenarios. We use open-source driving datasets in conjunction with Inverse Reinforcement Learning (IRL) to enhance the performance of our motion planner. We provide insight into our design choices and trade-offs made to achieve this solution. We also explore the impact of each component in the overall performance of our solution, with the intent of providing a guideline where allocation of resources can have the greatest impact.
\end{abstract}

\section{Introduction}

The CARLA Autonomous Driving (AD) challenge aims to advance autonomous vehicle research and development by focusing on their ability to excel in challenging traffic scenarios. The 2023 CARLA AD challenge adopts Leaderboard 2.0 as the evaluation framework, introducing a significant evolution from its predecessor, Leaderboard 1.0. Notably, Leaderboard 2.0 presents a heightened level of complexity by incorporating challenging scenarios, including open-door maneuvers, yielding to emergency vehicles, and more. Despite advances in end-to-end solutions and their superior performance, modular approaches provide abstractions that can yield much faster development. The inherent interpretability of modular solutions is another point of strength that differentiate them from end-to-end solutions. Another challenge associated with end-to-end solutions is their dependency on availability of good expert data. As of writing this report, unlike Leaderboard 1.0, Leaderboard 2.0 does not provide an autopilot as an expert driver for data collection. Consequently, training models exclusively on datasets from Leaderboard 1.0 proves insufficient for effectively addressing the challenges posed by Leaderboard 2.0 scenarios.

%As existing solutions \cite{LRM, chitta2022transfuser, shao2023reasonnet, TCP, chitta2021neat, jaeger2023hidden} primarily target Leaderboard 1.0 scenarios, a notable gap exists in solutions tailored specifically for the enhanced challenges of Leaderboard 2.0 scenarios. To address this gap, 

We present our Kyber-E2E solution, which secured the top rank in the 2023 CARLA AD challenge on the map track. We employ a modular approach which allows us to reuse components trained based on other datasets in the absence of expert data for Leaderboard 2.0. The key components of our solution includes sensing, localization, perception, tracking/prediction, and planning/control. In the realm of perception, we enhance object detection performance by integrating state-of-the-art language-assisted vision models. Leveraging these advanced models allows our solution to interpret complex scenes with heightened accuracy and efficiency. For tracking and prediction, our approach integrates the Unscented Kalman Filter (UKF) \cite{ukf} in conjunction with an unbalanced linear-sum assignment \cite{linear_sum_assignment} to effectively track and predict the trajectories of objects in dynamic environments. To fine-tune the parameters of our motion planner, we employ Inverse Reinforcement Learning (IRL) \cite{ratliff2009learning}. Particularly, we use IRL over InD open-source dataset \cite{inDdataset} to optimize the parameters of our planner. This synergistic approach facilitates the development of a robust and adaptable motion planner capable of navigating complex driving scenarios in a short amount of time. Our experimental results confirm the effectiveness of our planner, demonstrating its capability to navigate diverse and challenging driving scenarios presented in the CARLA AD challenge. 

The contributions of this paper is two-fold: 
\begin{itemize}
  \item Empirically show a modular design with components trained on different dataset is an effective approach. 
  \item Analyse the impact of each component on overall performance with the aim of showcasing where its better to allocate engineering and development resources.
\end{itemize}

In section II we present details about the design of each component and their development. Section III discusses the training and development of the models. We present the empirical result and analysis about impact of various modules in section IV. Conclusion and limitation of our work are provided in Section V.

\section{Agent Architecture}
\begin{figure*}
  \centering
  \includegraphics[width=0.85\textwidth]{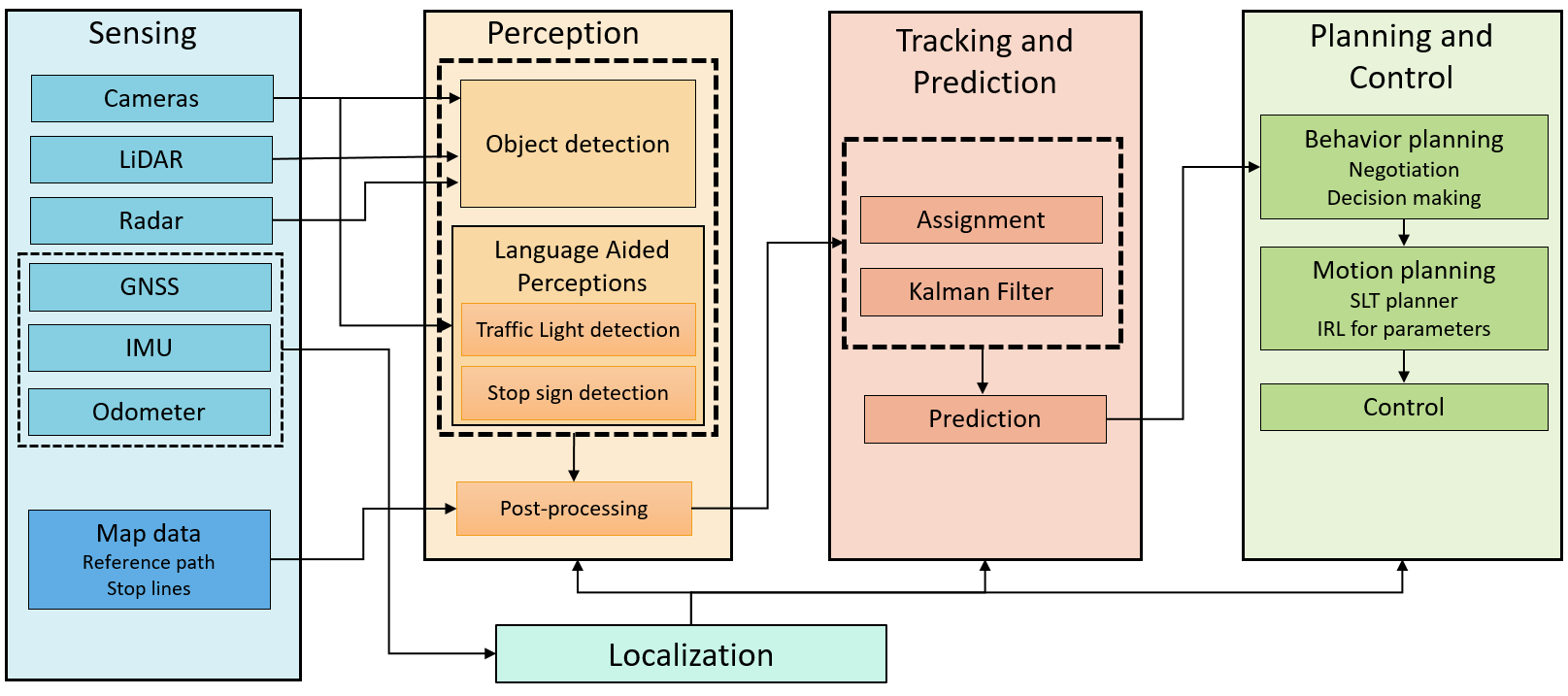}
  \caption{Kyber-E2E agent architecture.}
  \label{architecture}
\end{figure*}

The architecture of our agent is given in Fig. \ref{architecture}. In what follows, we discuss the main components of our architecture design. 

\subsection{Sensing}
Our sensing module consists of the following complementary sensors: 
\begin{itemize}
    \item One Front-facing RGB camera, used for object detection and traffic signal detection.  %on-road and one upward facing camera to detect traffic lights.
    \item One 360-deg LiDAR used for object detection.
    \item One radar to estimate velocity and position of far dynamic objects. 
    \item GNSS, IMU, and odometer to estimate ego vehicle state.
    \item OpenDrive map that is used to extract reference path and improve the perception output
\end{itemize}

%, namely, 360-deg LiDAR installed at 3D location $(0,0,2.5)$ m, and a single forward-looking front camera installed at 3D location $(-1.5,0,2)$ m, with rotation $R$, set to the $3\times3$ identity matrix. The size of the RGB image is $256$ rows by $1024$ rows, and the horizontal FOV is $110$ deg.

% We used a set of sensors to measure and collect raw data from the external environment and the vehicle itself. Fig. (-) shows the sensor layout of our agent, including two cameras, one radar, one LiDAR, IMU, GNSS, and speedometer.

%The sensor setup used is composed of the following:
%\begin{itemize}
%    \item for localization: The OpenDrive map, and a GNSS modules are used (default configuration).
%    \item for ego vehicle state: An IMU, and a Speedometer sensors are utilized (default configuration).
%    \item for Perception,  \emph{LAPS} \ref{laps} uses a front facing camera with a
%\end{itemize}

\subsection{Perception}

\begin{figure}
  \centering
  \includegraphics[width=0.48\textwidth]{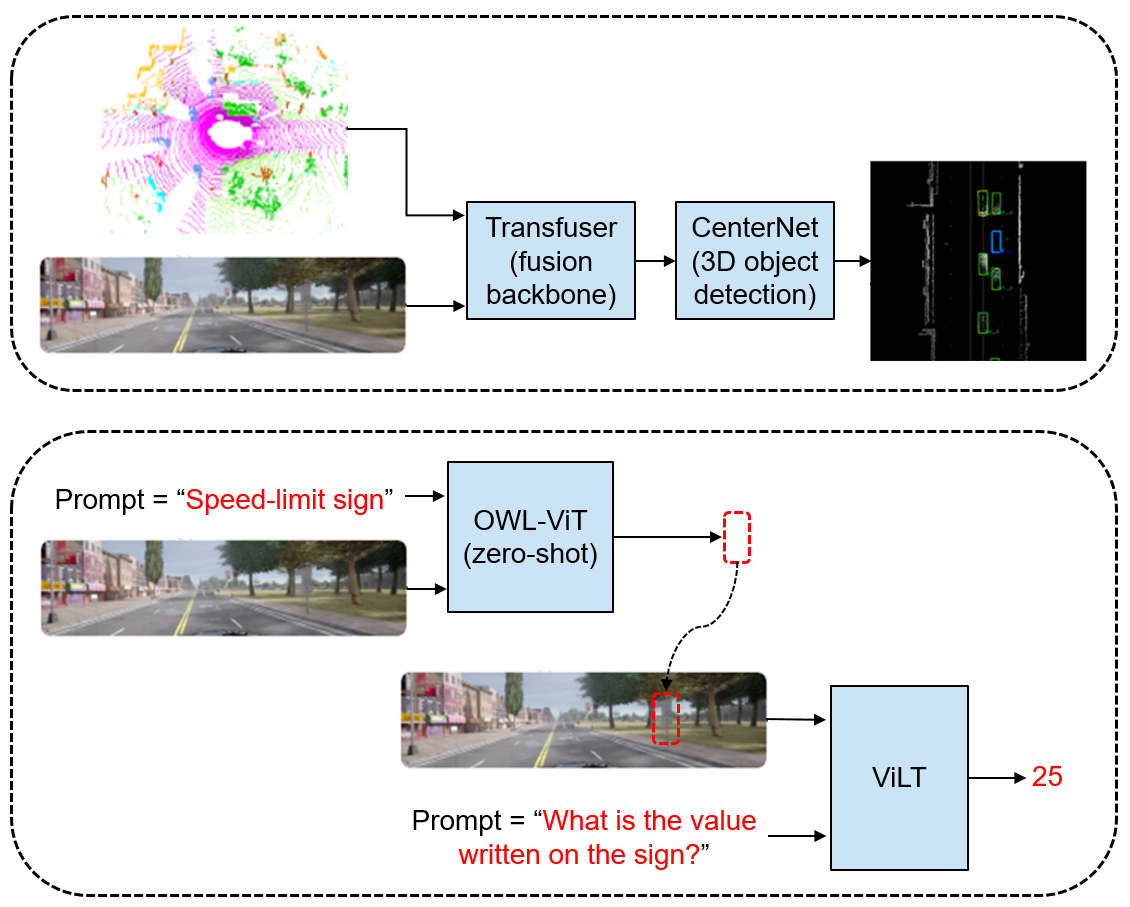}
  \caption{perception modules.}
  \label{perception}
\end{figure}

The perception module deals with instantaneous observation of the environment handling dynamic and static object detection and traffic signals. Temporal perceptual information is handled by tracking and prediction modules downstream to perception, which are discussed in \ref{tracking}. The architecture for perception is decomposed into two main components, as shown in Fig. \ref{perception}. The first component focuses on object detection and provide the location, orientation, and size of objects. The second one deals with traffic signs including traffic light, stop sign and speed limit signs. In what follows, we briefly describe the architecture of each module.%, and 3) a post-processing to make perception outputs consistent to map data. %backup module that uses the radar sensor and deals specifically with special detection cases.

\subsubsection{Signs Detection and Recognition} \label{laps}
This part of perception deals more exclusively with camera based perception. For this purpose we feed a high resolution (1080p) front facing camera feed to a combination of two pre-trained open-source models. The OWL-ViT \cite{minderer2022simple} model is used for Zero-shot First-person-view (FPV) 2D object detection and ViLT \cite{kim2021vilt} model is used for Visual Question Answering in the same view point. We find that using zero-shot detection is enough for the level of performance that was needed. The camera feed is first segmented into regions of interest based on assumptions about the potential location of each traffic sign. The segmented regions are then separately fed into OWL-ViT to detect the bounding boxes based on the newly defined zero-shot classes \emph{(e.g. "traffic light", "stop sign", "speed sign")}.
\\
Azimuth, altitude and distance are calculated using the detected bounding boxes and the intrinsic and extrinsic of the camera sensor. Bounding boxes are also used to further crop the segments of the image to check for additional inquiries about the signs. For this purpose, the cropped images are fed into ViLT along with language based queries (prompts) to get the traffic light state and the speed limit value that is written on the sign. The prompts used for the winning submission are \emph{("What color is the traffic light?", "what is the speed written on the sign?")}. We dub this sub-module the \textbf{\emph{Language Aided Perceptions (LAPS).}}

\subsubsection{Object Detection} \label{tfbox}
This module is a modified variant on top of the work from \cite{chitta2022transfuser}, using the backbone as described in Transfuser along with the Centernet head ~\cite{zhou2019objects}. The input to this branch is Birds-eye-view (BEV) voxelization of the lidar point cloud along with a front facing camera. It passes through the fusion backbone and the 3D object detection head. This outputs the position, orientation, size and class information of the detected objects. We augment the existing classes to include (cars, bikes, pedestrians, construction areas, cars with open door, and emergency vehicles).
\\
% The ground truth for some of these classes are not explicitly defined in CARLA such as the cars with open door so we add some logic to find the vehicles with their door open and extend their dimensions to include the door area. Another example is construction area where we find the construction sign/panel and the nearby cones and generate a box that encapsulates them together. %Some other tricks like fixing erroneous orientations and dimensions are also used but are left out of this report. %This submodule of perception is dubbed \textbf{\emph{TFBox}}. 
%Furthermore, to enrich the object detection, particularly for objects that are far away from the ego we use radar. In fact, it is executed parallel to the primary object detection process and extends the list of detected objects. We finetune the detections based on OpenDrive map to fix erroneous orientations.
The Lidar-based object detection has a limited range of 32 meters. To enrich object detection, especially for objects located far away from the ego, we employ radar. This process runs concurrently with the primary object detection, extending the list of identified objects. 

\subsubsection{Post-processing} \label{alignment}
The outputs from the two detection modules are further enhanced using the OpenDrive map. Based on certain rules and expert knowledge, we match objects and traffic signs with the maps elements to rectify any erroneous detections and orientations.

%It is worth mentioning that 
%\subsubsection{Radar Perception}
%for the majority of the competition scenarios, the deployed object detection is sufficient. However, making a maneuver involving invading the oncoming lane requires a certain level of safety guarantee. In this scenario, the ego vehicle is blocked by a construction area, while the other lane is in the opposite direction with constant traffic. The ego must find a gap in the oncoming traffic to make the bypass. To achieve this goal, we use a radar that returns a point cloud with altitude, azimuth, distance and radial velocity information. The points are filtered (to keep further objects with lower detection accuracy), a threshold is defined on the speed to perform the filtering. The points are then clustered using Kmeans \cite{1056489} and the clusters recognized as additional detection that is added onto the detection list from primary perception. 

%To improve the accuracy of our perception and make it robust to noises, we anchor objects' orientations to the map using OpenDrive map data.

%In addition, 
%\subsubsection{Post-processing}
%to make perception outputs robust to noises and to improve the accuracy of perception model, we anchor objects' orientations to the map. %Also, we use map information to assign conservative initial speed to objects. That is crucial to ensure our tracking and prediction module does not return unrealistic speed values. 

\subsection{Tracking and prediction} \label{tracking}
\subsubsection{Tracking}
\begin{figure}
  \centering
  \includegraphics[width=0.5\textwidth]{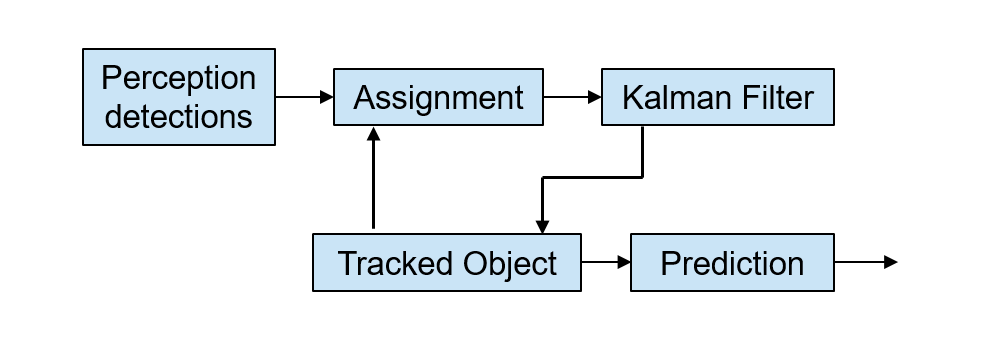}
  \caption{Tracking/prediction module}
  \label{prediction}
\end{figure}

Tracking is essential in our solution for two reasons. Given the zero shot perception we employed, there is no speed information for dynamic objects. The objects need to be tracked over time to estimate their speed for effective prediction and planning. Additionally, certain scenarios in CARLA Leaderboard 2.0 necessities tracking of object to estimate their behavior for proper decision making. The architecture of our tracking/prediction module is shown in Fig. \ref{prediction}. At every time-step, the perception module measures the ego-frame position and the orientation of all the objects around the ego. All the measurements are then converted to the global frame to make the measurements independent from the ego state. Moreover, we need to improve robustness of the perception measurements and determine the temporal relationship between consecutive perception measurements. To achieve this goal, we use 
%The perception measurements are not directly useful for planning due to two codependent issues: 1- The perception measurements are neither accurate nor robust. 2- The temporal relationship between consecutive perception measurements is not directly available. We propose to 
UKF alongside an unbalanced linear-sum assignment to track the objects along time, filter the noise, and improve robustness. %The measurement dimension of the Kalman Filter is set to 3.% (MEASUREMENT-DIM = 3). 
%We further define our state to be the position (X-Y), orientation (Yaw), and the speed (V). %Accordingly, our state dimension is 4. %(STATE-DIM = 4). 
%We use a simple constant velocity kinematic model as our state transition function. %\ref{eq:1}. 
%\begin{equation}
%    \begin{split}
%         X_{t+1} = X_{t} + Speed_{t} * cos(Yaw_{t}) \\
%         Y_{t+1} = Y_{t} + Speed_{t} * sin(Yaw_{t}) \\
%         Yaw_{t+1} = Yaw_{t} \\
%         Speed_{t+1} = Speed_{t}  \\
%    \end{split}
%    \label{eq:1}
%\end{equation}
We use unbalanced linear-sum assignment to assign the new perception observations to the already active tracks. The associated cost for each assignment is the norm of error between the UKF prediction and the candidate observation. If the minimum assignment cost for an active track is above a fixed maximum allowable cost, the track will not be assigned to any of the observations. If a track is not assigned to any observation for a fixed amount of simulation time-steps (MAX-ACTIVE-TIME), the track will be dropped. On the other hand, if an observation is not assigned to any of the tracks, we initiate a new track for that observation. To suppress noise, a new track will become active only after if it is assigned to an observation for a fixed amount of simulation time-steps (TIME-To-INIT). When an observation is assigned to a track, we use the observation to update the Kalman Filter state. We set the values of MAX-ACTIVE-TIME and TIME-To-INIT according to the object types observed by the perception module. For example, for pedestrians and bikes we set a high MAX-ACTIVE-TIME and a low TIME-To-INIT so that the planner is more conservative towards these minority yet important object classes.

\subsubsection{Prediction}
We use output of the UKF to predict other objects' future trajectories which is required by the planning module. To achieve this goal, we assume a constant speed along the anchored path if the object moves along a lane. Contrarily, if the object crosses a lane, both speed and heading are kept constant.

%For each object, the prediction module uses an angle threshold of 15 degrees to determine if it is moving along the nearest lane or crossing the nearest lane. For the former, its predicted path is along the center line of that lane. For the latter, its predicted path keeps the current heading in a straight line. In both cases, the predicted trajectory points are generated assuming constant speed. If an object moving along a lane meets a fork, the one with minimum heading change takes priority.

\subsection{Planning and Control}
Our planning module consists for three sub-modules: 1) The behavior planning makes the high level decisions, 2) the motion planner generates safe and feasible motions according to bheavior planning's output, and 3) controller convert the trajectory from motion planner to throttle/brake and steering command. 
\subsubsection{Behavior planning}
Behavior planning consists of negotiation and lane decision. The role of negotiation module is to calculate the assumed acceleration of each object. It assumes an internal model for each object and apply a heuristic rule over the prediction module's output to anticipate reaction of other objects to possible ego behavior. Objects following the ego or moving toward it are assumed to have a negative acceleration, while emergency vehicles driving toward the ego are assumed to have a positive acceleration. For other objects, interception points with the ego's reference path are calculated, determining the required ego acceleration to yield/unyield. After estimating the accelerations, the predicted trajectory for each object is then adjusted.

%To negotiate with A state machine is generated for each object to be negotiated with.

%, and a hysteresis of 1 exists for acceleration thresholds. All parameters are fine-tuned through trial and error. The predicted trajectory for each object is then adjusted based on the assumed acceleration, potentially becoming longer (accelerating) or shorter (decelerating).

%If an object is following the ego in the same lane, or its path is towards the ego body, it is assumes to have a negative acceleration of -5 \(m/s^2\). If it is an emergency vehicle (police car, ambulance and fire truck) driving towards ego, it is assumed to have a positive acceleration of 3 \(m/s^2\). 

%This approach filters out most of the noise from the prediction. The rest of the objects are to be negotiated with. For each object whose predicted path intercepts ego's reference path, the interception point is calculated, and the required ego's acceleration to yield/unyield to this object is calculated. If the threshold to unyield is lower than 2 \(m/s^2\), this object is supposed to yield to ego, and a negative acceleration of -3 \(m/s^2\) is assigned to it. Please note that a state machine is generated for each object to be negotiated with, and hysteresis of 1 \(m/s^2\) exist for acceleration threshold. All parameters are tuned by trial and error.

%Once the assumed acceleration is determined for each object, its predicted trajectory is then altered to fit for the acceleration. It could become longer (accelerating) or shorter (decelerating).

For the decision making, the first choice of the lane which the ego is following is the reference path. However, based on the position of the front object, there could be multiple options. If the front object is overlapping with the reference path, the ego is supposed to keep lane and follow it. If the front object is deviating to one side of the lane and leaving enough room on the other side (bicycles and pedestrians), the ego is supposed bypass it. If the front ego is stopped, or it is a construction site (road blocker), the ego is supposed to change lane. Note that in some cases, the other lane is in another direction. In such cases, the ego should find a long enough gap (40 \(m\)+5\(\times\)oncoming speed+road blocker length) in the oncoming traffic lane before initiating the lane change. Based on the target lane and the avoidance, the reference path is translated, and is passed to the motion planner.

\subsubsection{Motion Planning}
A sample-based SLT planner \cite{zhang2022spatial} is utilized for motion planning. Laterally, with the current heading as the initial heading, it samples $11$ Bezier curves as potential paths to account for the need to deviate laterally from the reference path. Longitudinally, with the current speed as initial speed, it samples $12$ speed profiles with constant accelerations. In total, $132$ potential trajectories are generated. These trajectories are evaluated using a combination of costs, including: 1) Swiftness cost, which is defined as the L2 difference integration of the acceleration and 3\(m/s^2\); 2) Longitudinal jerk cost, which is defined as the L2 integration of longitudinal jerk; 3) Lateral jerk cost, which is defined as the L2 integration of the lateral jerk; 4) Close to reference path cost, which is defined as the L2 difference integration of the trajectory and the reference path; and 5) Safety cost, which is defined as the $\text{exp}(-\gamma d)$, where $\gamma$ is the safety cost parameter and $d$ is the minimum polygon-to-polygon distance between ego's trajectory and object's prediction. 
Often the challenge in motion planning is finding the suitable set of weights to balance the various costs and achieve the desired behavior. We utilized Inverse Reinforcement Learning, specifically the Maximum Margin Planning algorithm from \cite{ratliff2009learning} to find the weights assigned to each cost and the parameter $\gamma$.

\subsubsection{Controller}
The controller is decoupled into lateral controller and longitudinal controller. They are both classic PID controllers. The lateral controller takes heading error as input, and steering angle as output. %Its PID coefficients are 0.9, 0.75, 0. 
The longitudinal controller takes acceleration error as input, acceleration command as intermediate result, and with calibrated longitudinal dynamics as feed-forward lookup table, throttle/brake output is generated. %Its PID coefficients are 5, 0.75, 1.0.

\section{Experiments and result}

\subsection{Data}
Most modules in our solution are either designed with hand or pre-trained models, with the exceptions being the object detection sub-module and the motion planner. 

In the absence of expert trajectory data for learning the motion planner module, we leveraged the inD datatset \cite{inDdataset} to tune our planner. Given the prominence of unsignalized intersections in leaderboared 2.0 scenarios, we found the inD dataset to be a suitable choice for this purpose.  

Given the need for identification of wider classes of objects in Leaderboard 2.0 scenarios, there was need to collect suitable for training the object detection submodule. We collected about 12 hours of new data from CARLA Town 12 according to the routes and scenarios provided as part of Leaderboard 2.0. For this purpose, we employed the developed prediction and planning modules and replaced the perception and tracking modules with privileged information directly received from CARLA simulator. The data was collected with randomly varying weather and lighting conditions in addition to augmentation of camera and lidar through rotation along z axis.

\subsection{Object Detection Training}

The training process for the perception model has three phases. First, we trained our model for $100$ epochs on a set of $40K$ frames captured from CARLA Town 12, and evaluated on $5337$ validation frames from Town 13.  We adopted the AP/mAP (Mean Average Precision) BEV object detection evaluation metrics implementation from ~\cite{openpcdet2020}. The trained model performed reasonably well on the car class. However, due to class imbalance issue, its performance was unfavorable for the rest of the classes. To address this issue, we collected a set of short sequences focused on capturing instances of the under-represented classes, and prepared a new training set of $42$K class-balanced training frames with global rotation augmentations to continue training for another $50$ epochs. Thanks to this, we observed significant improvements in the under-represented classes. %For our final $50$-epoch training for CARLA submission, we used a set of $56$K training frames captured from both Town 12 and Town 13. 
For all experiments, we used $8$ Nvidia Tesla V100 GPUs with batch size of $14$, with constant learning rate set to $0.0005$, and AdamW optimizer as done in ~\cite{jaeger2023hidden}. 

\section{Analysis and Result}
We evaluated our model on the set of validation routes from Town13 provided as part of Leaderboard 2.0 and present the evaluation result provided by the Leaderboard. The Leaderboard assesses agents based on their driving score ((DS), calculated as the product of route completion (RC) and infraction penalty (IS). Route completion measures the extent to which the agent follows the planned route, while the infraction penalty penalizes the agent for violations of traffic rules or collisions. For every violation in a route the infraction penalty is multiplied by certain fraction. A higher driving score indicates superior driving performance. Each number is the average across 20 routes from the Town13 validation routes.

CARLA Leaderboard 2.0 poses increased challenges compared to its predecessor, introducing new scenarios that demand agents to navigate complex situations, such as handling open doors, yielding to emergency vehicles, exiting parking, etc. As a result, scores from Leaderboard 1.0 are incomparable to those from Leaderboard 2.0.

\subsection{CARLA Leaderboar 2.0 Result}

The CARLA Leaderboard 2.0 for 2023 competition in the MAP Track is presented in in Table \ref{table_results_official}\footnote{It is reported by CARLA team that that more than 20 teams participated and made more than 100 submissions. However, only two teams made their submission public for the MAP Track.}.

Our experiments demonstrate that our planner adeptly manages numerous challenging scenarios, including yielding to emergency vehicles, lane changes, etc. However, as we integrate rules into our planner, its performance becomes more dependent on perception module's accuracy. An error in the perception module can lead the agent to come to a halt. For instance, when the agent needs to navigate into the oncoming lane to circumvent existing traffic, we require high-range information to determine if there is a sufficient gap for the ego vehicle to proceed. Since this high-range information is presently unavailable, the planner does not perform optimally in these situations.

\setlength{\tabcolsep}{2pt}
\renewcommand{\arraystretch}{1.3}
\begin{table*}[t]
\small
\caption{Evaluation on the impact of various modules on the overall performance of the AD solution}
\label{table_results}
\centering
\begin{tabular}{ p{3cm}|c c c|c c c c c c c c c c c c  }
\noalign{\hrule height 0.1cm}
\textbf{exp}& \textbf{DS} $\uparrow$ & \textbf{RC} $\uparrow$ & \textbf{IS} $\uparrow$ & \textbf{Ped} $\downarrow$ & \textbf{Veh} $\downarrow$ & \textbf{Stat} $\downarrow$ & \textbf{Red} $\downarrow$ & \textbf{Stop} $\downarrow$ & \textbf{Dev} $\downarrow$ & \textbf{Spd} $\downarrow$ & \textbf{Emrg} $\downarrow$ & \textbf{STO} $\downarrow$ & \textbf{Rdev} $\downarrow$ & \textbf{Block} $\downarrow$ & \textbf{RTO} $\downarrow$ \\
\hline
Mp              & 27.25 & 87.11 & 0.36 & 0.17 & 1.72 & 0.56 & 0.33 & 0.28 & 0.17 & 0.17 & 0.22 & 0.56 & 0.00 & 0.28 & 0.00\\
Mp+SensorEgoPos & 24.69 & 83.10 & 0.39 & 0.00 & 1.72 & 0.22 & 0.33 & 0.17 & 0.17 & 0.06 & 0.28 & 0.61 & 0.00 & 0.33 & 0.00\\
Mp+SensorSign   & 24.69 & 83.10 & 0.39 & 0.00 & 1.72 & 0.22 & 0.33 & 0.17 & 0.17 & 0.06 & 0.28 & 0.61 & 0.00 & 0.33 & 0.00\\
Mp+32m          & 7.76 & 76.15 & 0.20 & 0.06 & 3.88 & 0.18 & 0.41 & 0.06 & 0.47 & 0.06 & 0.29 & 1.41 & 0.00 & 0.41 & 0.00\\
Mp+Track        & 11.84 & 81.99 & 0.22 & 1.33 & 2.47 & 0.40 & 0.27 & 0.33 & 0.20 & 0.13 & 0.20 & 0.67 & 0.00 & 0.33 & 0.00\\
\hline
Ms+NoProcess    & 2.45 & 8.59 & 0.52 & 0.11 & 1.11 & 0.11 & 0.50 & 0.17 & 0.17 & 0.06 & 0.00 & 0.39 & 0.00 & 1.00 & 0.00\\
Ms+PrivEgoPos   & 1.46 & 35.62 & 0.14 & 1.72 & 5.44 & 2.00 & 1.39 & 1.00 & 0.33 & 0.33 & 0.00 & 0.94 & 0.00 & 0.78 & 0.06\\
Ms+PrivSign     & 6.17 & 30.69 & 0.29 & 0.56 & 3.39 & 0.67 & 0.33 & 0.22 & 0.39 & 0.06 & 0.00 & 0.50 & 0.00 & 1.00 & 0.00\\
\hline
Ms              & 1.99 & 39.01 & 0.14 & 0.72 & 4.11 & 0.94 & 0.72 & 0.39 & 0.28 & 0.22 & 0.00 & 1.22 & 0.00 & 0.94 & 0.00\\
\noalign{\hrule height 0.05cm}
\end{tabular}
\end{table*}

\subsection{Impact of Modules}
To assess the effectiveness of each module we performed a range of experiments were modules were replaced with their privileged counterparts. The result are summarized in table Table \ref{table_results}. On one end of spectrum we have the \textit{Privileged} agent (Mp) that utilizes the simulators privileged perception and tracking information. On the other end is the submitted solution (Ms) that does not use any privileged information.

\begin{table}[t]
\caption{CARLA official Leaderboard 2.0 - MAP Track - 2023}
\small
\label{table_results_official}
\centering
\begin{tabular}{c|c|c|c}
\noalign{\hrule height 0.05cm}
Team & Driving Score & Route Completion & Infraction Penalty \\
\hline
Kyber-E2E & 3.109 & 5.285 & 0.669 \\
\hline
LRM & 1.14 & 3.65 & 0.46 \\
\noalign{\hrule height 0.05cm}
\end{tabular}
\end{table}

\subsubsection{Localization}
We performed two experiments \textbf{Mp+SensorEgoPos} and \textbf{Ms+PrivEgoPos} to override the localization module with sensor-based and privileged localization respectively and assess the impact of localization. While there is variation due to randomness of the simulator, the replacing localization does not have a significant impact on the performance.

\subsubsection{Detection Range}
To simulate the impact of the limited range of lidar-based object detection, we performed an experiment \textbf{Mp+32m} where the privileged objects retrieved from CARLA simulator is limited to a 32 meter radius. We see a significant increase in vehicle collisions (\textbf{Veh}) and consequently drop in driving score (\textbf{DS}). These collisions are happening in three type of scenarios: 1) emergency braking in highway and suburban roads with high speed limit, 2) bypassing blocked lanes through incoming lane, and 3) unprotected left turn. These are all scenarios where the relative speed of ego and other vehicles are relatively high and the 32 meter does not provide enough time for ego to avoid a collision. 

\subsubsection{Traffic Signs}
The two experiments \textbf{Mp+SensorSign} and \textbf{Ms+PrivSign} override the LAPS module for traffic sign detection with sensor-based and privileged traffic sign detection, respectively. 
comparing the \textbf{Ms+PrivSign} and \textbf{Ms} we see a notable improvement in DS. Intuitively, this improvement can be attributed to the improved red light violation (\textbf{Red}), stop sign infraction (\textbf{Stop}), and minimum speed infraction (\textbf{Spd}) due to improved detection of traffic signs. Additionally, we see notable improvement in scenario time out (\textbf{STO}). This is due to the LAPS-based traffic sign detection identifying green lights as red light in harsh weather condition, causing ego to stop indefinitely behind a traffic light resulting in a scenario time out violation.  

Similarly, we see degraded performance when the privileged traffic sign information is replaced with sensor-based information when comparing  \textbf{Mp} and  \textbf{Mp+SensorSign}

\subsubsection{Tracking}
We conduct an experiment \textbf{Mp+Track} to evaluate the effect of the tracking module. Here the IDs provided by the simulator is removed and the tracking module tries to track objects and assign appropriate ID to them. Compared to \textbf{Mp}, there is notable impact on IS which in turn result in reduce DS. The infraction with the most notable increase is the collision with pedestrians (Ped). This is likely a limitation of our tracking module as it was mainly tuned for vehicles and cannot handle pedestrians properly. 

\subsubsection{Post-processing}
The post-processing module takes the detected object and refine the result based on map information. We performed an experiment \textbf{Ms+NoProcess} where we remove the post-processing step. The intuitive expectation is to get reduced DS; however, we see DS is higher for this experiment compared to \textbf{Ms}. We noticed that the agent with no post-processing often gets blocked due to erroneous detection and result in an agent blocked (Block) result. This is evident in RC metric, whcih is significantly lower in this experiment. Due to lower driving distance, the number of infractions are also much lower resulting in better IS value compared to \textbf{Ms}. This is a byproduct of the way CARLA's score metrics are designed. When the number of infractions are relatively high, between two agents that make the same number of infractions per kilometer, the agent that drives further will likely get lower DS. For example if an agent drives 10\% and makes one infraction with penalty of 0.5 the total driving score will be 5. Another agent that drives 40\% and makes 4 infractions with each having a penalty of 0.5 will have driving score of 2.5. Given the complexity of Leaderboard 2.0 and limited performance of our module the number of infractions are relatively high. Hence supposedly lower performing model is getting a higher DS. 

%Table \ref{table_results} presents the performance of our agent in CARLA Leaderboard 2.0 challenge, map track. The main metric of the Leaderboard is driving score which is defined as the product between the route completion and the infractions penalty. 

%It is worth mentioning that the results of Leaderboard 2.0 are significantly different from previous Leaderboard 1.0 due to several reasons. First, the challenging scenarios such as open-door, yielding to emergency vehicles, exiting parking, etc. has been added to this challenge which makes the evaluation more challenging. Moreover, for participating in Leaderboard 1.0 scenarios, there ia an autopilot which can be used as expert for data collection. In other words, one difficulty of Leaderboard 2.0 comes from the fact that we do not have data collection agent. 

%No limit infraction as our sign detection module 

\section{Conclusion and Limitations}
We introduced Kyber-E2E solution that secured the top spot in the 2023 CARLA AD challenge, map track. Our five-component architecture, encompassing sensing, localization, perception, tracking/prediction, and planning/control, proved effective in surpassing the challenges posed by the evolved Leaderboard 2.0. the emperical result show that in a modular architecture with right abstraction, modules developed independetly with different dataset can still yield reasonably well performance. While we achieved the top spot in Leaderboard 2.0 in 2023 competition, we acknowledge the dependence of our planner on accurate perception, particularly in highly-crowded scenes. The need for high-range information, especially in situations requiring lane changes into oncoming traffic, presents an avenue for future refinement. We anticipate addressing these challenges through the implementation of a fully end-to-end autonomous driving architecture in our future work.

\bibliographystyle{unsrt}
\bibliography{references}

\end{document}